\documentclass[conference]{IEEEtran}
\IEEEoverridecommandlockouts
\usepackage{cite}
\usepackage{amsmath,amssymb,amsfonts}
\usepackage{algorithmic}
\usepackage{graphicx}
\usepackage{textcomp}
\usepackage{xcolor}

\usepackage{pgfplots}
\pgfplotsset{compat=1.17} 
\usepackage{tikz}
\usepackage{adjustbox}
\usepackage{amsmath}
\usepackage{booktabs}
\usepackage{multirow}
\usepackage{tabularx}
\usepackage{url}

\def\BibTeX{{\rm B\kern-.05em{\sc i\kern-.025em b}\kern-.08em
    T\kern-.1667em\lower.7ex\hbox{E}\kern-.125emX}}
\begin{document}

\title{WBT-BGRL: A Non-Contrastive Weighted Bipartite Link Prediction Model for Inductive Learning
\thanks{This study was financed in part by the Coordenação de Aperfeiçoamento de Pessoal de Nível Superior – Brasil (CAPES) – Finance Code 001. We thank CNPq and FAPESP Grant 2020/09850-0 for financial support. }

}

\author{
\IEEEauthorblockN{
    Joel Frank Huarayo Quispe\IEEEauthorrefmark{1}, \ 
    Lilian Berton\IEEEauthorrefmark{2}, \  and
    Didier Vega-Oliveros\IEEEauthorrefmark{3}
}
\IEEEauthorblockA{
    \textit{Institute of Science and Technology}\\
    \textit{Federal University of São Paulo (UNIFESP)}\\
    São José dos Campos, Brazil \\
    \{%
    \IEEEauthorrefmark{1}joel.frank,
    \IEEEauthorrefmark{2}lberton,
    \IEEEauthorrefmark{3}didier.vega%
    \}@unifesp.br
}}


\maketitle



\begin{abstract}
Link prediction in bipartite graphs is crucial for applications like recommendation systems and failure detection, yet it is less studied than in monopartite graphs. Contrastive methods struggle with inefficient and biased negative sampling, while non-contrastive approaches rely solely on positive samples. Existing models perform well in transductive settings, but their effectiveness in inductive, weighted, and bipartite scenarios remains untested. To address this, we propose Weighted Bipartite Triplet–Bootstrapped Graph Latents (WBT-BGRL), a non-contrastive framework that enhances bootstrapped learning with a novel weighting mechanism in the triplet loss. Using a bipartite architecture with dual GCN encoders, WBT-BGRL is evaluated against adapted state-of-the-art models  (T-BGRL, BGRL, GBT, CCA-SSG). Results on real-world datasets (\textit{Industry} and \textit{E-commerce})  show competitive performance, especially when weighting is applied during pretraining—highlighting the value of weighted, non-contrastive learning for inductive link prediction in bipartite graphs.

\end{abstract}


\section{Introduction}

Graph Neural Networks (GNNs) are fundamental for processing graph-structured data and are widely applied in recommendation systems, social networks, and bioinformatics \cite{Khemani2024}. Among their key applications, link prediction is essential for uncovering new relationships~\cite{vega2021link} and has significantly advanced with the development of GNNs \cite{DBLP:journals/corr/abs-1802-09691}. However, this task presents major challenges in inductive settings, where models must generalize to unseen nodes or links, an essential requirement for real-world applications such as recommending new products \cite{shiao2023linkprediction}. These challenges are amplified in bipartite graphs (e.g., user-product networks), which are inherently sparse and common across many domains \cite{Zeng_2013}.

Traditionally, GNNs have been trained in a supervised fashion using labeled data. In contrast, Graph Self-Supervised Learning (Graph SSL) offers a robust alternative by pretraining models through internal tasks derived from the graph’s structure and node features, eliminating the need for manual labels. Within Graph SSL, contrastive methods, while effective, incur high computational costs and exhibit sensitivity due to extensive negative sampling, particularly in large, sparse bipartite graphs \cite{huang2025doesgclneedlarge, hu2024graphrankingcontrastivelearning, ali2024overfittingrobustnessquantityquality}. This limitation becomes more pronounced in inductive scenarios, where the burden of negative sampling grows as models must generalize to new nodes. For this reason, non-contrastive methods, focused solely on reinforcing similarity among connected nodes, emerge as a more scalable and efficient solution for sparse and dynamic contexts.

Despite their potential, the performance of non-contrastive methods in inductive link prediction on bipartite graphs remains underexplored, especially regarding the critical role of edge weighting (e.g., interaction frequency), which is intrinsic to many of these domains.

This work addresses this gap by proposing the Weighted Bipartite Triplet – Bootstrapped Graph Latents (WBT-BGRL) model. WBT-BGRL is a novel extension of T-BGRL \cite{shiao2023linkprediction}, specifically designed for inductive link prediction on weighted bipartite graphs. It explicitly incorporates edge weights via weight-aware graph augmentations and provides four configurable versions to examine the impact of weighting in both the pretraining and link prediction loss functions. This modularity allows us to investigate how explicit edge weight integration improves generalization and how it compares against unweighted adaptations of state-of-the-art models (T-BGRL \cite{shiao2023linkprediction}, BGRL \cite{thakoor2023largescalerepresentationlearninggraphs}, GBT \cite{Bielak_2022}, CCA-SSG \cite{zhang2021canonicalcorrelationanalysisselfsupervised}).

Our contributions are as follows: (1) Development of WBT-BGRL, a new model for weighted bipartite graphs; (2) Adaptation of existing non-contrastive architectures to the bipartite setting; (3) Comparative evaluation of the four WBT-BGRL variants against baseline models on real-world weighted bipartite datasets (Industry, E-commerce) under an inductive scenario; and (4) Analysis of how edge weighting affects model generalization during both pretraining and link prediction phases.

\section{Methodology}
In this section details the methodological design for link prediction in weighted bipartite graphs using a non-contrastive approach.
The proposed model architecture is presented, along with the adaptations made to the base models, and evaluation metrics.

\subsection{Proposed Model}
WBT-BGRL is an extension of the T-BGRL framework, adapted to operate on weighted bipartite graphs and to explore the impact of weighting on self-supervised learning and link prediction.
The architecture of WBT-BGRL is composed of the following main components:

\textbf{Encoder (WeightedGCNEncoder):}
    The encoder is a heterogeneous graph neural network (GNN) based on \texttt{GCNConv} layers from PyTorch Geometric, specifically designed for bipartite graphs with two types of nodes: $U$ (first partition) and $V$ (second partition). 
    Unlike standard encoders, this component accepts and incorporates edge weights $w_{ij}$ during message passing when weighted pretraining is enabled (WP variants). 
    The message passing at layer $\ell$ is expressed as:
    \begin{equation}
        H^{(\ell+1)} = \mathrm{ReLU}(\tilde{A}_{\text{weighted}} H^{(\ell)} W^{(\ell)}),
    \end{equation}
    where $H^{(\ell)}$ are the concatenated node embeddings at layer $\ell$, 
    $W^{(\ell)}$ is the trainable weight matrix, and 
    $\tilde{A}_{\text{weighted}}$ is the symmetrically normalized adjacency matrix that includes edge weights $w_{ij}$ when active.
    The encoder outputs separate embeddings for each node type: 
    $H_U$ and $H_V$.

 \textbf{Projectors:}
    Two independent two-layer MLPs transform the encoder outputs into projection spaces for the self-supervised objective:
    \begin{align}
        Z_U &= W_2^U \cdot \text{PReLU}(W_1^U H_U + b_1^U) + b_2^U, \\
        Z_V &= W_2^V \cdot \text{PReLU}(W_1^V H_V + b_1^V) + b_2^V,
    \end{align}
    where $Z_U$ and $Z_V$ are the projected embeddings for node types $U$ and $V$.

 \textbf{Predictors (Key Architectural Difference):}
    Unlike T-BGRL which uses a shared predictor, WBT-BGRL introduces two independent predictors one for each node type to capture bipartite asymmetries:
    \begin{align}
        P_U &= W_2^{P_U} \cdot \text{PReLU}(W_1^{P_U} Z_U + b_1^{P_U}) + b_2^{P_U}, \\
        P_V &= W_2^{P_V} \cdot \text{PReLU}(W_1^{P_V} Z_V + b_1^{P_V}) + b_2^{P_V}.
    \end{align}
    This design enables distinct transformation strategies for each node type.

 \textbf{Target Network:}
    A momentum-updated copy of the online encoder, projectors, and predictors.
    Its parameters are updated via Exponential Moving Average (EMA):
    \begin{equation}
        \theta_{\text{target}} \leftarrow \tau \theta_{\text{target}} + (1 - \tau) \theta_{\text{online}},
    \end{equation}
    where $\tau = 0.99$. This stabilizes training by providing consistent targets.

 \textbf{Decoder (Link Predictor):}
    The decoder is a three-layer MLP that takes the concatenated frozen embeddings of two nodes (one from $U$, one from $V$)
    and predicts the probability of a link between them:
    \begin{equation}
        p(u,v) = \sigma(\mathrm{MLP}([h_u \,||\, h_v])),
    \end{equation}
    where $\sigma(\cdot)$ is the sigmoid function. 
    The decoder is trained using binary cross-entropy loss, optionally weighted (WB variants):
    \begin{equation}
        \mathcal{L}_{\text{decoder}} = -\frac{1}{|\mathcal{E}|} 
        \sum_{(u,v) \in \mathcal{E}} 
        [y \log p(u,v) + (1 - y) \log(1 - p(u,v))].
    \end{equation}
    Positive edges can be upweighted to mitigate class imbalance.
    
    \begin{itemize}
        \item {WBT-BGRL (WP, WB):} Both the pretraining loss and the link prediction loss use the actual edge weights.
        \item {WBT-BGRL (WP, NWB):} Only the pretraining loss incorporates the edge weights; the link prediction loss assumes uniform weights of 1.0.
        \item {WBT-BGRL (NWP, WB):} Only the link prediction loss uses the actual edge weights; the pretraining loss ignores them.
        \item {WBT-BGRL (NWP, NWB):} Neither loss uses real edge weights; both are computed as if all edge weights were 1.0 (equivalent to T-BGRL (Adapted)).
    \end{itemize}
    
    This systematic ablation allows identifying:
    
    \begin{itemize}
      \item Whether the weights add value in self-supervised pretraining
      \item Whether the weights add value in supervised link prediction
      \item Whether the benefit is additive (WP\_WB better than WP\_NWB and NWP\_WB)
    \end{itemize}

\subsubsection{Pre-training process}
This phase trains the \texttt{GCNEncoder} and the \texttt{Predictor} in a self-supervised manner.

{\bf Augmented View Generation:} Two augmented views of the training graph are generated using augmentation strategies designed to preserve structural information while introducing variability: 
\[
(\tilde{X}^{(1)}, \tilde{A}^{(1)}, \tilde{W}^{(1)}) \quad \text{and} \quad (\tilde{X}^{(2)}, \tilde{A}^{(2)}, \tilde{W}^{(2)})
\]
using the following strategies:

\begin{itemize}
    \item {Feature Dropping:} Randomly sets a percentage of node features in \( \tilde{X} \) to zero (default \( p = 0.1 \)).
    
    \item {Weight-Aware Edge Dropping:} Removes edges such that the probability of retaining an edge is proportional to its normalized weight \( w_{uv} \).
\end{itemize}

{\bf Corrupted View Generation:} The corrupted view is generated to provide negative examples that the model must learn to reject. This view deliberately disrupts the semantic structure of the graph.

\[
(\tilde{X}^{(c)}, \tilde{A}^{(c)}, \tilde{W}^{(c)})
\]
A corrupted view is generated via:

\begin{itemize}
    \item {Feature Shuffling:} Node features are randomly permuted. That is, a node $U$ now has the features of another node \( U' \neq U \) selected at random.

    \item {Random Bipartite Edges:} New edges are created randomly between partitions $U$ and $V'$, and assigned a minimum weight (typically 1.0).
\end{itemize}

\textbf{Weighted Pretraining Loss Calculation:}  
The objective is to make the embeddings of the same node (from different augmented views) similar, while making the embeddings from the corrupted view dissimilar. The \texttt{GCNEncoder} (online and target) processes the different views, and the \texttt{Predictor} \( \text{pred}(\cdot) \) operates on the output of the online encoder. The total pretraining loss \( \mathcal{L}_{\text{WBT-BGRL}} \) is defined as:

\begin{equation}
\mathcal{L}_{\text{WBT-BGRL}} = \lambda \cdot \mathcal{L}_{\text{repulsive}} + (1 - \lambda) \cdot \mathcal{L}_{\text{attractive}}
\end{equation}

where \( \lambda \) is a balancing hyperparameter (typically \( \lambda = 0.5 \)). The attractive and repulsive terms are weighted averages of cosine similarity over the edge sets of the augmented and corrupted views, 
respectively:

\footnotesize
\begin{equation}
\mathcal{L}_{\text{attractive}} = - \frac{1}{\sum_{(u,v) \in \tilde{\mathcal{E}}^{(1)}} w_{uv}^{(1)}} \sum_{(u,v) \in \tilde{\mathcal{E}}^{(1)}} w_{uv}^{(1)} \cdot \cos\left(\text{pred}(h_u^{(1)}), h_v^{(2)}\right)
\end{equation}


\footnotesize
\begin{equation}
\mathcal{L}_{\text{repulsive}} = \frac{1}{\sum_{(u,v) \in \tilde{\mathcal{E}}^{(c)}} w_{uv}^{(c)}} \sum_{(u,v) \in \tilde{\mathcal{E}}^{(c)}} w_{uv}^{(c)} \cdot \cos\left(\text{pred}(h_u^{(1)}), h_v^{(c)}\right)
\end{equation}
\normalsize


Here, \( h_u^{(1)} \) is the embedding of node \( u \) from the first augmented view (online encoder), \( h_v^{(2)} \) is the embedding of node \( v \) from the second augmented view (target encoder), and \( h_v^{(c)} \) is the embedding from the corrupted view (target encoder). The sets \( \tilde{\mathcal{E}}^{(1)} \) and \( \tilde{\mathcal{E}}^{(c)} \) denote the edge sets of the augmented and corrupted views, respectively, with their corresponding weights \( w_{uv}^{(1)} \) and \( w_{uv}^{(c)} \).

\textbf{EMA Update:}  
The target encoder is updated via an Exponential Moving Average (EMA) of the online encoder’s weights to provide stability during training.

\subsubsection{Link Predictor Training}
Once pretraining is complete, the encoder is frozen (its weights are not updated) and a supervised decoder is trained for the link prediction task.

\begin{itemize}
    \item {Fixed Embeddings:} For each node in the validation set, embeddings are generated using the frozen encoder.
    Handling new nodes (inductivity):
    
    \begin{itemize}
      \item Known nodes: Use embeddings from the pretrained encoder.
      \item New nodes: Mapped to a special \texttt{<UNK>} token whose embedding is learned during pretraining.
    \end{itemize}

    \item {Negative Sampling for LP:} Negative links are randomly generated such that they respect the bipartite structure and do not exist in the original graph. These are used for both training and evaluation of the Link Predictor.
    
    \item {Weighted BCEWithLogitsLoss:} The PyTorch \texttt{BCEWithLogitsLoss} function is used, which combines a sigmoid function \( \sigma(\cdot) \) and binary cross-entropy loss. The loss is explicitly weighted by the edge weights \( w_j \), and defined as:
    
 \footnotesize   
    \begin{equation}
        \mathcal{L}_{\text{LP}} = -\frac{1}{M} \sum_{j=1}^{M} w_j \left[y_j \log \sigma(\hat{s}_j) + (1 - y_j) \log (1 - \sigma(\hat{s}_j))\right]
    \end{equation}
\normalsize
    where \( M \) is the total number of links (positive and negative) in the batch, \( w_j \) is the weight assigned to link \( j \) (typically the real edge weight for positives, and 1.0 for negatives), \( y_j \) is the true label (1 if the link exists, 0 otherwise), and \( \hat{s}_j \) is the predicted logit (pre-sigmoid score) produced by the Link Predictor.
    
    This weighting ensures that the model prioritizes learning from more frequent or important edges.
\end{itemize}

\subsection{Non-contrastive models adapted to bipartite structures}

To evaluate WBT-BGRL, we compare it with four state-of-the-art non-contrastive self-supervised models for bipartite graphs: T-BGRL, BGRL, GBT, and CCA-SSG \cite{shiao2023linkprediction}. Originally designed for unweighted homogeneous graphs, all were adapted to heterogeneous bipartite graphs with two node types.
Main Adaptations to Bipartite Graphs:

\begin{enumerate}
    \item \textbf{Dual Architecture:} Two separate encoders (\texttt{u\_encoder}, \texttt{v\_encoder}) for heterogeneous node types with independent latent spaces ($Z_u \in \mathbb{R}^{N_u \times d}$, $Z_v \in \mathbb{R}^{N_v \times d}$).
    
    \item \textbf{Asymmetric Projections:} Cross-type projections (\texttt{u$\rightarrow$v\_proj}, \texttt{v$\rightarrow$u\_proj}) align heterogeneous embeddings for loss computation.
    
    \item \textbf{Directed Message Passing:} Bipartite-preserving aggregation ($U \rightarrow V$, $V \rightarrow U$ only) instead of homogeneous ($N \rightarrow N$).
    
    \item \textbf{Multi-Component Loss:} Expanded from 1 to 2--4 components to capture both intra-type consistency and cross-type relationships:
    \begin{itemize}
        \item \textbf{T-BGRL / BGRL:} $L = L_u + L_v$ (bootstrap on both types)
        \item \textbf{GBT:} $L = L_{uu} + L_{vv} + L_{uv} + L_{vu}$ (four correlation matrices)
        \item \textbf{CCA-SSG:} $L = L_{\text{cca}} + \lambda_u \cdot \text{decorr}_u + \lambda_v \cdot \text{decorr}_v$ (dual constraints)
    \end{itemize}
    
    \item \textbf{Dual Target Networks (T-BGRL / BGRL):} Independent EMA updates for each node type, allowing asynchronous learning rates.
    
    \item \textbf{Type-Specific Augmentations:} Differentiated augmentation strategies while preserving bipartite structure (no $U \leftrightarrow U$ or $V \leftrightarrow V$ edges created).
\end{enumerate}

All adaptations maintain self-supervised learning principles while handling $U \leftrightarrow V$ heterogeneity through architectural duplication and asymmetric operations.

\subsection{Evaluation Metrics}
The experiments were conducted on Google Colab using an NVIDIA T4 GPU. Each model was run 5 times, reporting average performance and standard deviation. Link prediction was evaluated using AUC-ROC, AP, Hits@50, precision, recall, and F1 score (threshold = 0.5).

\section{Experiments}

\subsection{Experimental Setup for Inductive-Temporal Evaluation and Datasets}
To evaluate the model, we used two datasets (one public and one private). Node representations were enriched with categorical and numerical features compatible with heterogeneous GNN architectures.

A strictly chronological and inductive data split was applied, dividing the dataset into 80\% for training (\texttt{train\_df}), 10\% for validation (\texttt{val\_df}), and 10\% for testing (\texttt{test\_df}). This temporal split ensures a realistic inductive evaluation~\cite{vega2019evaluating} setting, where link prediction is performed on future nodes and interactions not observed during encoder training.

The encoder was trained exclusively on the \texttt{train\_df} graph. The decoder was trained on positive links from \texttt{val\_df}, using embeddings produced by the frozen encoder applied to \texttt{train\_df}. Final evaluation was performed using embeddings computed from the full history available up to the test period (\texttt{train\_df + val\_df}) to predict the links in \texttt{test\_df}. Negative samples for both validation and testing were generated by sampling $U \leftrightarrow V$ pairs not present in any part of the dataset (\texttt{train\_df}, \texttt{val\_df}, or \texttt{test\_df}), maintaining the inductive nature of the setup.

The full pipeline was executed five times using different random seeds (42 to 46) to account for variability introduced. Link prediction performance was measured using \textbf{ROC-AUC}, \textbf{AP}, \textbf{F1}, \textbf{Recall}, \textbf{Precision} (at threshold 0.5), and \textbf{Hits@50}, with results reported as \textbf{mean ± standard deviation across 5 runs}.

\begin{itemize}
   \item \textbf{Industry:} Bipartite graph with node types \texttt{[locations, assets]}.
    \begin{itemize}
        \item \textbf{Node U (locations):} Represented by a learnable embedding derived from its unique location ID.
        \item \textbf{Node V (assets):} Includes a categorical attribute \texttt{equipment\_type}, encoded via a learnable embedding.
    \end{itemize}

    \item \textbf{E-commerce:} Bipartite graph with node types \texttt{[customers, products]}.
    \begin{itemize}
        \item \textbf{Node U (customers):}
        \begin{itemize}
            \item \texttt{scaled\_mean\_spend}: Log1p-transformed and MinMax-scaled average spend.
            \item \texttt{country\_code}: Integer-encoded country, used as an embedding index.
        \end{itemize}
        \item \textbf{Node V (products):}
        \begin{itemize}
            \item \texttt{scaled\_avg\_price}: Log1p-transformed and MinMax-scaled average price.
            \item \texttt{popularity}: Number of unique purchasing customers, used as an embedding index.
        \end{itemize}
    \end{itemize}

    \item \textbf{Graph Weighting:}
    \begin{itemize}
        \item \textit{Unweighted}: Used for all four baseline models.
        \item \textit{Weighted (by edge frequency)}: Used for WBT-BGRL and all its four variants, applied to both datasets.
    \end{itemize}
\end{itemize}

\subsection{Experimental Setup}

To ensure the robustness of the results, each experiment was run five times with different random initializations. We report the mean and standard deviation of the evaluation metrics.

We compare against four state-of-the-art self-supervised models, which we adapt to the bipartite setting: T-BGRL, BGRL, GBT, and CCA-SSG (all adapted bipartite versions).
Additionally, we introduce \textbf{WBT-BGRL}, with four variants: \texttt{WP\_WB} (weighted pretraining + weighted decoder), \texttt{WP\_NWB}, \texttt{NWP\_WB}, and \texttt{NWP\_NWB} (no weights), where edge weights reflect transaction frequencies.


\textbf{Training Protocol.} 
We employ a two-phase approach: 
\begin{enumerate}
    \item \textbf{Phase 1:} The encoder is trained in a self-supervised manner on \textsc{Past} data for 200 epochs. After training, all encoder parameters (including embeddings) are frozen.
    \item \textbf{Phase 2:} A 2-layer MLP decoder is trained in a supervised fashion on \textsc{Present} data for 100 epochs, with early stopping based on Hits@50 (patience = 10).
\end{enumerate}
The final evaluation measures temporal link prediction performance on \textsc{Future} data.

\textbf{Hyperparameters.} 
All models use the following shared hyperparameters: \texttt{hidden\_dim = 256}, \texttt{output\_dim = 128}, \texttt{num\_layers = 2}, \texttt{dropout = 0.2}, \texttt{lr = 0.001}, \texttt{weight\_decay = 1e-5}, and \texttt{batch\_size = 512}. 
Model-specific parameters (e.g., $\tau = 0.99$ for EMA, $\lambda = 5 \times 10^{-3}$ for CCA-SSG).\\

\textbf{The code used in this study  is available at:}\\ \url{https://github.com/JoelFrank/bipartite-temporal-link-prediction}






\section{Results}
This section presents a comprehensive evaluation of non-contrastive self-supervised learning methods for temporal link prediction on two real-world bipartite graphs.


\begin{table*}[t]
\centering
\caption{LINK PREDICTION RESULTS ON ALL DATASETS (MEAN ± STANDARD DEVIATION OVER \textbf{5 RUNS}). WBT-BGRL CONFIGURATIONS: \textbf{WP} = WEIGHTED PRETRAIN LOSS, \textbf{NWP} = NON-WEIGHTED PRETRAIN; \textbf{WB} = WEIGHTED BCE LOSS, \textbf{NWB} = NON-WEIGHTED BCE.}
\label{tab:all_datasets}
\renewcommand{\arraystretch}{1.1}
\setlength{\tabcolsep}{4pt}
\begin{tabular}{|l|c|c|c|c|c|c|}
\hline
\multicolumn{7}{|c|}{\textbf{Industry}} \\
\hline
\textbf{Model} & \textbf{ROC-AUC} & \textbf{AP} & \textbf{F1-Score (t = 0.5)} & \textbf{Recall (t = 0.5)} & \textbf{Precision (t = 0.5)} & \textbf{Hits@50} \\
\hline
WBT-BGRL (WP, WB)   & 0.9161 ± 0.0038 & 0.9195 ± 0.0039 & 0.7865 ± 0.0016 & 0.9949 ± 0.0014 & 0.6503 ± 0.0028 & 0.6231 ± 0.0282 \\
WBT-BGRL (WP, NWB)  & 0.9742 ± 0.0053 & 0.9737 ± 0.0065 & 0.8988 ± 0.0187 & 0.8844 ± 0.0536 & 0.9169 ± 0.0345 & 0.8559 ± 0.0358 \\
WBT-BGRL (NWP, WB)  & 0.9848 ± 0.0023 & 0.9825 ± 0.0035 & 0.9137 ± 0.0108 & \textbf{0.9964 ± 0.0028} & 0.8438 ± 0.0199 & 0.9276 ± 0.0155 \\
WBT-BGRL (NWP, NWB) & \textbf{0.9944 ± 0.0025} & \textbf{0.9919 ± 0.0039} & \textbf{0.9855 ± 0.0030} & 0.9907 ± 0.0029 & \textbf{0.9803 ± 0.0060} & \textbf{0.9979 ± 0.0007} \\
\hline
T-BGRL (Adapted)     & 0.9995 ± 0.0004 & 0.9994 ± 0.0006 & 0.9950 ± 0.0006 & 0.9939 ± 0.0009 & 0.9962 ± 0.0014 & 0.9981 ± 0.0005 \\
BGRL (Adapted)       & 0.9997 ± 0.0004 & 0.9996 ± 0.0006 & \textbf{0.9971 ± 0.0009} & 0.9958 ± 0.0025 & \textbf{0.9985 ± 0.0009} & \textbf{1.0000 ± 0.0000} \\
GBT (Adapted)        & 0.9996 ± 0.0004 & 0.9994 ± 0.0007 & 0.9916 ± 0.0016 & \textbf{0.9966 ± 0.0005} & 0.9866 ± 0.0034 & \textbf{1.0000 ± 0.0000} \\
CCA-SSG (Adapted)    & \textbf{0.9998 ± 0.0002} & \textbf{0.9997 ± 0.0002} & 0.9899 ± 0.0070 & 0.9824 ± 0.0140 & 0.9976 ± 0.0018 & \textbf{1.0000 ± 0.0000} \\
\hline
\multicolumn{7}{|c|}{\textbf{E-commerce}} \\
\hline
\textbf{Model} & \textbf{ROC-AUC} & \textbf{AP} & \textbf{F1-Score (t = 0.5)} & \textbf{Recall (t = 0.5)} & \textbf{Precision (t = 0.5)} & \textbf{Hits@50} \\
\hline
WBT-BGRL (WP, WB)   & 0.8560 ± 0.0025 & 0.8279 ± 0.0024 & 0.7917 ± 0.0030 & 0.9557 ± 0.0054 & 0.6758 ± 0.0062 & 0.9705 ± 0.0021 \\
WBT-BGRL (WP, NWB)  & 0.8568 ± 0.0028 & 0.8287 ± 0.0025 & \textbf{0.7979 ± 0.0026} & 0.9120 ± 0.0110 & 0.7093 ± 0.0083 & 0.9708 ± 0.0023 \\
WBT-BGRL (NWP, WB)  & 0.8560 ± 0.0019 & 0.8282 ± 0.0021 & 0.7908 ± 0.0036 & \textbf{0.9584 ± 0.0106} & 0.6733 ± 0.0098 & 0.9709 ± 0.0015 \\
WBT-BGRL (NWP, NWB) & \textbf{0.8571 ± 0.0024} & \textbf{0.8292 ± 0.0027} & 0.7974 ± 0.0027 & 0.9181 ± 0.0166 & \textbf{0.7050 ± 0.0125} & \textbf{0.9710 ± 0.0016} \\
\hline
T-BGRL (Adapted)     & 0.8547 ± 0.0118 & \textbf{0.8309 ± 0.0067} & 0.7638 ± 0.0261 & 0.7632 ± 0.0633 & 0.7678 ± 0.0113 & 0.0332 ± 0.0039 \\
BGRL (Adapted)       & 0.8552 ± 0.0048 & 0.8262 ± 0.0049 & \textbf{0.7933 ± 0.0072} & \textbf{0.8765 ± 0.0277} & 0.7251 ± 0.0091 & 0.0255 ± 0.0015 \\
GBT (Adapted)        & 0.8395 ± 0.0102 & 0.8262 ± 0.0080 & 0.7232 ± 0.0211 & 0.6668 ± 0.0397 & \textbf{0.7918 ± 0.0071} & \textbf{0.0364 ± 0.0040} \\
CCA-SSG (Adapted)    & \textbf{0.8496 ± 0.0127} & 0.8250 ± 0.0084 & 0.7662 ± 0.0163 & 0.7749 ± 0.0284 & 0.7583 ± 0.0127 & 0.0272 ± 0.0032 \\
\hline
\end{tabular}
\end{table*}


\subsection{Results on the Industry Dataset}
The Industrial dataset shows a clear performance hierarchy (Table \ref{tab:all_datasets}, upper section). 

\textbf{Baseline Models:} achieve near-perfect performance: \textit{CCA-SSG} (ROC-AUC: $0.9998 \pm 0.0002$), \textit{BGRL} (F1: $0.9971 \pm 0.0009$, Hits@50: 1.0), \textit{GBT} (Recall: $0.9966 \pm 0.0005$), and \textit{T-BGRL} (ROC-AUC: $0.9995 \pm 0.0004$). All report ROC-AUC $>$ 0.999 with std $<$ 0.001.

\textbf{WBT-BGRL Variants:} show configuration sensitivity: \textit{NWP\_NWB} achieves best WBT-BGRL performance (ROC-AUC: $0.9944 \pm 0.0025$, F1: $0.9855 \pm 0.0030$, Hits@50: $0.9979 \pm 0.0007$), followed by \textit{NWP\_WB} ($0.9848 \pm 0.0023$). \textit{Weighted pretraining (WP)} variants drop significantly: \textit{WP\_NWB} ($0.9742 \pm 0.0053$, $-2\%$) and \textit{WP\_WB} ($0.9161 \pm 0.0038$, $-7.8\%$).

\textbf{Impact of Weighting:} The dataset's skewed edge frequencies (max = 377) cause \textit{WP\_WB} overfitting ($-7.8\%$ vs. \textit{NWP\_NWB}). Removing weighted BCE (\textit{WP\_NWB}) partially recovers performance ($-2\%$), but \textit{NWP} variants' uniform treatment during pretraining yields superior generalization.


\subsection{Results on the E-commerce Dataset}
E-Commerce reveals different dynamics with modest performance and higher variance (Table \ref{tab:all_datasets}, lower section).

\textbf{WBT-BGRL Performance:} All variants achieve competitive, tightly clustered results: \textit{NWP\_NWB} (ROC-AUC: $0.8571 \pm 0.0024$, Precision: $0.7050 \pm 0.0125$, Hits@50: $0.9710 \pm 0.0016$), \textit{WP\_NWB} (F1: $0.7979 \pm 0.0026$, best), and \textit{WB} variants (Recall: $\approx 0.957$). All four configurations span ROC-AUC $0.8560$–$0.8571$ ($< 0.15\%$ difference), showing minimal weighting impact due to balanced frequencies (max = 58) and high new-edge ratio (77.7\%).

\textbf{Baseline Performance:} \textit{CCA-SSG} (ROC-AUC: $0.8496 \pm 0.0127$), \textit{BGRL} (F1: $0.7933 \pm 0.0072$, Recall: $0.8765 \pm 0.0277$), \textit{T-BGRL} (AP: $0.8309 \pm 0.0067$), and \textit{GBT} (Precision: $0.7918 \pm 0.0071$) show comparable global metrics but dramatically low Hits@50 ($0.0255$–$0.0364$, 2.5–3.6\%).

\textbf{Hits@K Gap:} WBT-BGRL's Hits@50 ($\approx 97\%$) vs. baselines ($\approx 3\%$) reflects architectural trade-offs: WBT-BGRL's frozen-encoder + decoder training concentrates predictions in top-K, while baselines' end-to-end optimization prioritizes global discrimination (AP, ROC-AUC) over top-K precision.

\section{Conclusions}
This study evaluated WBT-BGRL against four adapted baselines on inductive-temporal link prediction, revealing dataset-specific trade-offs:

\begin{itemize}
    \item \textbf{Dataset-Specific Patterns:} Baselines dominated Industry (ROC-AUC $>$ 0.999, Hits@50 = 1.0) by 0.5–8\% over WBT-BGRL (best: \textit{NWP\_NWB} = $0.9944$, worst: \textit{WP\_WB} = $0.9161$). In E-Commerce, WBT-BGRL achieved 97\% Hits@50 vs. baselines' 3\% ($+94\%$) while matching global metrics (ROC-AUC $\approx$ 0.857 vs. 0.850–0.855).
    
    \item \textbf{Edge Weighting Impact:} \textit{NWP} outperformed \textit{WP} in skewed distributions (Industry: $-7.8\%$ drop from extreme outlier, max = 377), but  minimal effect in balanced data (E-Commerce: $< 0.15\%$ variance, max = 58). Weighting benefits require moderate variance without outliers.
    
    \item \textbf{Architectural Trade-off:} Baselines excel at global discrimination (Industry ROC-AUC $>$ 0.999, E-Commerce AP = 0.831); WBT-BGRL at top-K concentration (Hits@50 = 97\% vs. 3\%). Use baselines for high-recall tasks (maintenance, anomaly detection), WBT-BGRL for top-K recommendations.
    
    \item \textbf{Design Recommendations:} (1) Use \textit{NWP} by default; \textit{WP} only if frequency std $< 5$ and no outliers. (2) Frozen-encoder for top-K (K $\leq$ 50), end-to-end for global ranking. (3) Independent predictors per node type capture bipartite asymmetries.
\end{itemize}


\textbf{Future work:}
Explore adaptive edge weighting schemes, hybrid models combining top-K and global strengths, and extensions to dynamic graphs.


\bibliographystyle{IEEEtran}
\bibliography{bibliography}
\vspace{12pt}

\end{document}